\pdfoutput=1

\documentclass[11pt]{article}

\usepackage{acl}

\usepackage{times}
\usepackage{latexsym}

\usepackage[T1]{fontenc}

\usepackage[utf8]{inputenc}

\usepackage{microtype}

\usepackage{inconsolata}
\usepackage{booktabs}
\usepackage{multirow}
\usepackage{amsmath}
\usepackage{amsfonts}
\usepackage{cleveref}
\crefformat{section}{\S#2#1#3}
\crefformat{subsection}{\S#2#1#3}
\crefformat{subsubsection}{\S#2#1#3}
\crefformat{appendix}{Appendix #2#1#3}
\crefformat{equation}{Eq.~(#2#1#3)}
\crefformat{table}{Table~#2#1#3}
\crefformat{figure}{Figure~#2#1#3}

\title{The Devil is in the Details: On Models and Training Regimes \\ for Few-Shot Intent Classification}

\author{
Mohsen Mesgar$^1$ 
\And
Thy Thy Tran$^1$ 
\And
Goran Glava{\v s}$^2$ \\ 
\And
Iryna Gurevych$^1$ 
\AND 
$^1$Ubiquitous Knowledge Processing Lab (UKP) \\ 
Department of Computer Science \\
Technical University of Darmstadt \\
$^2$CAIDAS, University of Würzburg, Germany \\
\url{www.ukp.tu-darmstadt.de}
}

\usepackage{tikz}
\usetikzlibrary{matrix}
\usepackage{xspace}
\newcommand{\NEL}{NE\xspace}
\newcommand{\EL}{EP\xspace}
\newcommand{\ELSQ}{EPSQ\xspace}
\newcommand{\BE}{BE\xspace}
\newcommand{\CE}{CE\xspace}
\newcommand{\Pa}{PA\xspace}
\newcommand{\Me}{NP\xspace}

\begin{document}
\maketitle

\begin{abstract}
Few-shot Intent Classification (FSIC) is one of the key challenges in modular task-oriented dialog systems. 
While advanced FSIC methods are similar in using pretrained language models to encode texts and nearest-neighbour-based inference for classification, these methods differ in details. 
They start from different pretrained text encoders, use different encoding architectures with varying similarity functions, and adopt different training regimes.
Coupling these mostly independent design decisions and the lack of accompanying ablation studies are big obstacle to identify the factors that drive the reported FSIC performance. 
We study these details across three key dimensions: 
(1) \textit{Encoding architectures}: Cross-Encoder vs Bi-Encoders;
(2) \textit{Similarity function}: Parameterized (i.e., trainable) functions vs non-parameterized function;  
(3) \textit{Training regimes}:  Episodic meta-learning vs the straightforward (i.e., non-episodic) training. 
Our experimental results on seven FSIC benchmarks reveal three important findings. 
First, the unexplored combination of the cross-encoder architecture (with parameterized similarity scoring function) and episodic meta-learning consistently yields the best FSIC performance. 
Second, Episodic training yields a more robust FSIC classifier than non-episodic one.
Third, in meta-learning methods, splitting an episode to support and query sets is not a must. 
Our findings paves the way for conducting state-of-the-art research in FSIC and more importantly raise the community's attention to details of FSIC methods. 
We release our code and data publicly. 
\end{abstract}

\section{Introduction}
\label{sec:intro}

Intent classification deals with assigning one label from the predefined set of classes or \textit{intents} to user utterances. 
This task is vital for understanding user demands in dialogue systems and the predicted intent of an utterance is a key input to other modules (i.e., dialog management) in  task-orientated dialog systems \cite{ma-etal-2022-effectiveness,louvan2020recent,razumovskaia2021crossing}.
Although this task has been widely studied, the task is still challenging when dialogue systems including their intent classifiers should be extended to a wide variety of domains.
One of main challenges in training intent classifiers is costly labelled utterances~\cite{zhang-etal-2022-fine,wen2017network,budzianowski2018multiwoz,rastogi2020towards,hung2022multi2woz,mueller-etal-2022-label}. 
Therefore, the ability to adjust intent classifiers to new intents, given only a few labelled instances is imperative. 

Various methods (\cref{sec:rel}) for few-shot intent classification (FSIC) have been proposed  \cite{larson2019evaluation,casanueva2020efficient,zhang-etal-2020-discriminative,MehriDialoGLUE2020,krone-etal-2020-learning,casanueva-etal-2020-efficient,nguyen2020dynamic,zhang-etal-2021-shot,dopierre-etal-2021-protaugment,vulic-etal-2021-convfit,zhang2022fine}.  
These method are generally similar in utilizing pretrained language models (LMs) and resorting to $k$ nearest neighbour ($k$NN) inference -- the label of a new instance is determined based on the labels of instances with which it has the highest similarity score. 
Despite these similarities, these FSIC methods differ in detailed but shared crucial design dimensions including encoding architectures, similarity functions, and training regimes. 
These methods couple different choices across these dimensions, hindering ablations and insights into which factors drive the performance. 

In this work, we propose a formulation for comparing nearest neighbour-based FSIC methods (\cref{sec:method}). 
Within this scope, our formulation focuses on three key design decisions:
\textbf{(1)} \textit{model architecture for encoding utterance pairs}, where we contrast the less frequently adopted Cross-Encoder architecture (e.g., \cite{vulic-etal-2021-convfit}) against the more common Bi-Encoder architecture \cite{zhang-etal-2020-discriminative,krone-etal-2020-learning,zhang-etal-2021-shot}\footnote{Also known as Dual Encoder or Siamese Network.};
\textbf{(2)} \textit{similarity function} for scoring utterance pairs based on their joint or separate representations, contrasting the parameterized (i.e., trainable) neural scoring components against cosine similarity as the simple non-parameterized scoring function; 
and 
\textbf{(3)} \textit{training regimes}, comparing the standard non-episodic training (adopted, e.g., by \newcite{zhang-etal-2021-shot} or \newcite{vulic-etal-2021-convfit}) against the episodic meta-learning training (implemented, e.g., by \newcite{nguyen2020dynamic} or \newcite{krone-etal-2020-learning}).

We use our formulation to conduct empirical multi-dimensional comparison for two different text encoders (BERT \cite{devlin-etal-2019-bert} as a vanilla PLM and SimCSE \cite{gao-etal-2021-simcse} as the state-of-the-art sentence encoder) and, more importantly, under the same evaluation setup (datasets, intent splits, evaluation protocols and measures) while controlling for confounding factors that impede direct comparison between existing FSIC methods. 
Our extensive experimental results reveal two important findings. 
First, a Cross-Encoder coupled with episodic training, which has never been previously explored for FSIC, consistently yields best performance across seven established intent classification datasets. 
Second, although episodic meta-learning methods split utterances of an episode into a support and query set during training, for the first time, we show that this is not a must. 
In fact, without such splits the FSIC methods generalize better than (or similar to) the case without such splits to unseen intents in new domains.

In sum, our work raises the attention of the community to the importance of the pragmatical details, which are formulated as three dimensions, in the performance achieved by recent FSIC methods.
Alongside our novel findings pave the way for future research in conducting comprehensive FSIC methods. 
\section{Related Work}
\label{sec:rel}
Our work focuses on few-shot intent classification (FSIC) methods, which use the nearest neighbor ($k$NN) algorithm. 
Therefore, we describe existing inference algorithms and why we focus on $k$NN-based methods. 
Then, we categorize the literature about $k$NN-based methods concerning our three evaluation dimensions. 

\paragraph{Inference algorithms for FSIC.} 
Classical methods \cite{Xu2013ConvolutionalNN,Meng2017DialogueIC,Wang2019DialogueIC,Gupta2019SimpleFA} for FSIC use the maximum likelihood algorithm, where a vector representation of an utterance is given to a  probability distribution function to obtain the likelihood of each intent class. 
Training such probability distribution functions, in particular when they are modeled by neural networks, mostly requires a large number of utterances annotated with intent labels, which are substantially expensive to collect for any new domain and intent.
With advances in pre-trained language models, recent FSIC methods leverage the knowledge encoded in such language models to alleviate the need for training a probability distribution for FSIC. 
These advanced FSIC methods \cite{krone-etal-2020-learning,casanueva-etal-2020-efficient,nguyen2020dynamic,zhang-etal-2021-shot,dopierre-etal-2021-protaugment,vulic-etal-2021-convfit,zhang2022fine} mostly use the nearest neighbor algorithm ($k$NN-based) to find the most similar instance from a few labeled utterances while classifying an unlabelled utterance. 
These methods then identify the label of the found utterance as the intent class of the unlabelled utterance. 
Since nearest neighbor-based FSIC methods achieve state-of-the-art FSIC performance, we focus on the major differences between these methods as our comparison dimensions. 

\paragraph{Model architectures for encoding utterance pairs.}
One of the main differences between the $k$NN-based FSIC methods is their model architecture for encoding two utterances. 
The dominant FSIC methods \cite{zhang-etal-2020-discriminative,krone-etal-2020-learning,zhang-etal-2021-shot,Xia2021PseudoSN} use Bi-Encoder architecture \cite{NIPS1993_288cc0ff,reimers2019sentence,zhang-etal-2022-fine}. 
The core idea of Bi-Encoders is to map an unlabled utterance and a candidate labeled utterance separately into a common dense vector space and perform similarity scoring via a distance metric such as dot product or cosine similarity. 
In contrast, some FSIC methods \cite{vulic-etal-2021-convfit,zhang-etal-2020-discriminative,Wang2021MeLLLE,zhang-etal-2021-shot} use the Cross-Encoder architecture \cite{devlin-etal-2019-bert}.  
The idea is to represent a pair of utterances together using an LM, where each utterance becomes a context for the other. 
A Cross-Encoder does not produce an utterance embedding but represents the semantic relations between its input utterances. 
In general, Bi-Encoders are more computationally efficient than Cross-Encoders because of the Bi-Encoder's ability to cache the representations of the candidates.
In return, Cross-Encoders capture semantic relations between utterances where such relations are crucial for nearest neighbour-based FSIC methods.

\paragraph{Similarity scoring function.}
A crucial component in nearest neighbor-based methods for FSIC is the employed similarity function. 
This function estimates the similarity between input utterances to LMs. 
Concerning this comparison dimension, we categorize FSIC methods into two groups. 
First, FSIC methods \cite{zhou-etal-2022-knn,zhang-etal-2020-discriminative,Xia2021PseudoSN} which use parametric neural layers to estimate the similarity score between utterances. 
Second, those methods \cite{sauer-etal-2022-knowledge,zhang-etal-2022-fine,krone-etal-2020-learning,vulic-etal-2021-convfit,zhang2022fine,xu21_interspeech,zhang-etal-2021-shot}, which  rely on non-parametric methods (a.k.a metric-based methods) such dot-product, cosine similarity, and Euclidean distance function.

\paragraph{Training strategy.}
To simulate FSIC, the best practice is to split an intent classification corpus into two disjoint sets of intent classes.  
In this way, one set includes high-resource intents to train a FSIC classifier, and the other set includes low-resource intents to evaluate the classifier. 
Concerning the training strategy on the high-resource intents, FSIC methods can be divided into two clusters. 
One cluster of methods \cite{zhang-etal-2022-fine,nguyen2020dynamic,krone-etal-2020-learning} adopts meta-learning or episodic training. 
Under this training regime, the goal is to train a meta-learner that could be used to quickly adapt to any few-shot intent classification task with very few labeled examples. 
To do so, the set of high-resource intents are split to construct many episodes, where each episode is a few-shot intent classification for a small number of intents. 
The other cluster includes methods \cite{zhang-etal-2021-shot,vulic-etal-2021-convfit,xu21_interspeech,Xia2021PseudoSN,zhang-etal-2020-discriminative,zhang-etal-2021-shot} that use conventional supervised  (or non-episodic) training.  
The non-episode training strategy takes into the set of high-resource intents as a large training set and fine-tune the parameters of the FSIC model on all samples in this set.

\section{Method}
\label{sec:method}

We first describe the commonly adopted FSIC framework based on utterance similarities and nearest neighbour inference algorithm. 
We then present the alternative configurations along our three central dimensions of comparison: (i) model architecture for encoding utterance pairs, (ii) functions for scoring utterance pairs, and (iii) training regimes.    

\subsection{Nearest Neighbours (NN) inference.}
Following previous work on FSIC \cite{zhang-etal-2020-discriminative,vulic-etal-2021-convfit}, we cast the FSIC task as a sentence similarity task in which each intent being a latent semantic class that is represented by sentences associated with the intent.
The task is then to find the most similar labelled utterances for the given query/input that can directly derive the underlying semantic intent.
During inference, an FSIC method should deal with an $N$-way $k$-shot intent classification.  
$N$ is the number of intents, and $k$ is the number of labeled utterances given for each intent label. 
The advanced FSIC methods infer the intent of an utterance (i.e., query) based on its similarity with a given few labeled utterances. 

Let $q$ be a query utterance and $C = \{ c_1, ..., c_n \}$ be a set of its labeled neighbours. 
The nearest neighbour inference relies on a similarity function, non-parameterized or trainable (which is learned on high-resource intents), to estimate the similarity score $s_i$ between the query utterance $q$ and any neighbour $c_i \in C$. 
The query's label $\hat{y}_q$ is inferred as the ground-truth label of the neighbour with the maximum similarity score (i.e., $k=1$ in $k$NN inference): $\hat{y} = y_k\text{, } k=\text{argmax}(\{ s_1,..., s_n \})\text{.}$ 

\subsection{Model Architectures for Encoding Utterance Pairs}
The main component of an FSIC model is an encoder which represents a pair of utterances: a query and a labelled utterance.  
We explore two model architectures as used in recent FSIC methods: \text{Bi-Encoder} and \text{Cross-Encoder}. 

\paragraph{Bi-Encoder (BE).} 
BE encodes a pair of utterances independently, deriving independent representations of the query and the labelled utterance.  
In particular, for each utterance $x$ in a pair, we pass, ``\mbox{\texttt{[CLS]} $x$}'', to a BERT-like language model and use the vector representation of ``\texttt{[CLS]}'' to represent $x$.  
It is worth noting that the parameters of the LM are shared in BE.

\paragraph{Cross-Encoder (\CE).} 
Different from BE, CE encodes a pair of query $q$ and utterance $c_i$ \textit{jointly}.
We concatenate $q$ with each of its neighbours to form a set of query–neighbour pairs $P = \{ (q,c_1),...,(q,c_n) \}$.
We then pass each pair from $P$ as a sequence of tokens to a language model, which is pre-trained to represent the semantic relation between utterances. 
More formally, we feed a pair of utterances, ``\mbox{\texttt{[CLS]} $q$ \texttt{[SEP]} $c_i$}'', to a BERT-like LM and then use the representation of the ``\texttt{[CLS]}'' token as the representation of the pair.

\subsection{Similarity Scoring Function}
Given the pair representation, we can compute the similarity between a query and a labelled utterance by a \textit{parameterized} or \textit{non-parameterized} function. 

\paragraph{PArameterized (\Pa).} 
A neural-based parametric scoring function consists of a fully connected feed-forward network (FFN) that transforms the pair representation into a score
\begin{equation}
    s_i = \sigma \left( \mathbf{W}^{1\times d} \mathbf{h}_{(q, c_i)} + b \right) \text{,}
\end{equation} 
where the weight $\mathbf{W}$ and bias $b$ are trainable parameters, $d$ is the size of the representation $\mathbf{h}_{(q, c_i)} \in \mathbb{R}^d$, and  $\sigma(.)$ denotes the {\fontfamily{qcr}\selectfont sigmoid} activation function. 

\paragraph{Non-Parameterized (\Me).} 
In contrast to \Pa, \Me often uses vector-based similarity metrics as scoring functions, e.g., cosine similarity or Euclidean distance. 
Following \newcite{vulic-etal-2021-convfit}, in this work we adopt the cosine similarity between $h_q$ and $h_{c_i}$. 

\subsection{Model Configurations}

\begin{figure}[!t]
    \centering
    \resizebox{!}{0.10\textwidth}
    {
    \begin{tabular}{@{}c|c|c@{}}
\begin{tikzpicture}
    \node[] (c) at (0,0) {$c$};
    \node[] (q) at (1.5,0) {$q$};
    \node[draw, thick, minimum height = 1cm, minimum width =2cm] (lm) at (0.75,1.5) {\textsc{lm}};
    \draw[->, thick] (c) -- (0, 1.0);
    \draw[->, thick] (q) -- (1.5, 1.0);
    \node[draw, thick, minimum height = 1cm, minimum width =2cm] (mlp) at (0.75,3.0) {\textsc{ff}};
    \draw[->, thick] (lm) -- (mlp);
    \node[] (s) at (0.75,4.5) {$s$};
    \draw[->, thick] (mlp) -- (s);
\end{tikzpicture}
&
\begin{tikzpicture}
    \node[] (c) at (0.0,0) {$c$};
    \node[draw, thick, minimum height = 1cm, minimum width =2cm] (lm1) at (0.0,1.5) {\textsc{lm}};
    \draw[->, thick] (c) -- (lm1);

    \node[] (q) at (2.5,0) {$q$};
    \node[draw, thick, minimum height = 1cm, minimum width =2cm] (lm2) at (2.5,1.5) {\textsc{lm}};
    \draw[->, thick] (q) -- (lm2);
    
    \node[draw, thick, minimum height = 1cm, minimum width =2cm] (mlp) at (1.25,3.0) {\textsc{ff}};
    \draw[->, thick] (lm1) -- (mlp);
    \draw[->, thick] (lm2) -- (mlp);
    
    \node[] (s) at (1.25,4.5) {$s$};
    \draw[->, thick] (mlp) -- (s);
\end{tikzpicture}
& 
\begin{tikzpicture}
    \node[] (c) at (0.0,0) {$c$};
    \node[draw, thick, minimum height = 1cm, minimum width =2cm] (lm1) at (0.0,1.5) {\textsc{lm}};
    \draw[->, thick] (c) -- (lm1);

    \node[] (q) at (2.5,0) {$q$};
    \node[draw, thick, minimum height = 1cm, minimum width =2cm] (lm2) at (2.5,1.5) {\textsc{lm}};
    \draw[->, thick] (q) -- (lm2);
    
    \node[] (dotProduct) at (1.25,3.0) {$\bigodot$};
    \draw[->, thick] (lm1) -- (dotProduct);
    \draw[->, thick] (lm2) -- (dotProduct);
    
    \node[] (s) at (1.25,4.5) {$s$};
    \draw[->, thick] (dotProduct) -- (s);
\end{tikzpicture}
\\
\CE+\Pa 
&
\BE+\Pa
& 
\BE+NP
\end{tabular}
    }
    \caption{A demonstration of possible model configurations of encoder architectures and similarity functions to estimate the similarity score $s$  between a query $q$ and neighbour $c$.  
    CE and BE show Cross-Encoder and Bi-Encoder architectures using a BERT-like language model (LM), respectively. 
    PA and NP show parametric and non-parametric similarity functions, respectively. 
    PA is modeled by feedforward (FF) layers and NP by the dot product $\odot$. 
    }
    \label{fig:models}
\end{figure}
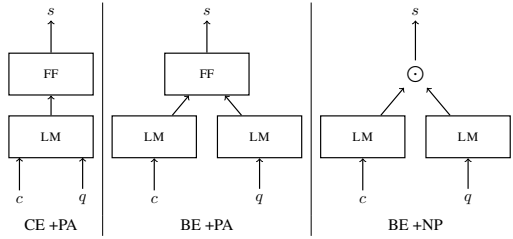

\cref{fig:models} illustrates the three possible combinations of encoders and similarity functions: (i) CE+\Pa; (2) BE+\Pa, and (3) BE+\Me.

\paragraph{\CE+\Pa.} 
In this configuration, we feed the joint encoding of the utterance pair to a parameterized similarity scoring function. 
We note again, due to a single representation vector for both utterances, \CE cannot be couples with a non-parameterized scoring (\Me). 

\paragraph{\BE+\Pa.}
In this configuration, we represent the pair by concatenating the representations of each utterance with the vectors of difference and element-wise product between those representations:
\begin{equation*}
   h_{(q, c_i)} = \hspace{-0.2em}(h_q \oplus h_{c_i} \oplus \lvert h_{q} - h_{c_i} \rvert \oplus h_{q} h_{c_i} ),
\end{equation*}
, where $\oplus$ is the concatenation operation.
We motivate the concatenation by the findings in \newcite{reimers-gurevych-2019-sentence}.  
Similar to \CE+ \Pa, we use the sigmoid activation function on top of the feed-forward layer. 
The size of $\mathbf{W}$ is then $1\times 4d $. 

\paragraph{\BE+\Me.} 
We use cosine similarity to estimate the similarity between input utterances during prediction. 
During training, we compute the dot product between the query and each neighbour representation vector to directly estimate the similarity scores $s_i =  \sigma \left(  h_q \odot h_{c_i} \right)$, where $\odot $ indicates the dot product, and $\sigma$ is the $sigmoid$ function. 
We apply $\sigma$ to scale $s_i$ to a value between $0$ and $1$. 

\begin{figure}[!t]
    \centering
    \resizebox{!}{0.48\textwidth}
    {
    \begin{tikzpicture}
    [
        box/.style={rectangle,draw=black, minimum size=1mm},
    ]
\matrix[matrix of nodes, nodes={draw,minimum size=1mm}, nodes in empty cells,column sep=-\pgflinewidth,row sep=-\pgflinewidth] (ne) at (0,0)
{
|[fill=red]| & |[fill=red]|  &  |[fill=red]|   &  |[fill=red]|  &  |[fill=red]|  &  |[fill=red]|  &  |[fill=red]| & |[fill=red]|  \\
|[fill=blue]| & |[fill=blue]|  &  |[fill=blue]|   &  |[fill=blue]|  &  |[fill=blue]|  & 
|[fill=blue]|  &  |[fill=blue]| & |[fill=blue]| \\
|[fill=green]| & |[fill=green]|  &  |[fill=green]|   &  |[fill=green]|  &  |[fill=green]|  &  |[fill=green]|  &  |[fill=green]| & |[fill=green]| \\
|[fill=gray]|  &  |[fill=gray]| & |[fill=gray]| & |[fill=gray]|  &  |[fill=gray]| & |[fill=gray]| &|[fill=gray]|  &  |[fill=gray]|  \\
|[fill=cyan]| & |[fill=cyan]|  &  |[fill=cyan]|   &  |[fill=cyan]|  &  |[fill=cyan]|  &  |[fill=cyan]|  &  |[fill=cyan]| & |[fill=cyan]| \\
|[fill=brown]| & |[fill=brown]|  &  |[fill=brown]|   &  |[fill=brown]|  &  |[fill=brown]|  &  |[fill=brown]|  &  |[fill=brown]| & |[fill=brown]| \\
};

\matrix[matrix of nodes, nodes={draw,minimum size=1mm}, nodes in empty cells,column sep=-\pgflinewidth,row sep=-\pgflinewidth] (ne) at (5,0)
{
|[fill=red]| & |[fill=red]|  &  |[fill=red]|   &  |[fill=white]|  &  |[fill=red]|  &  |[fill=red]|  &  |[fill=red]| & |[fill=red]|  \\
|[fill=blue]| & |[fill=blue]|  &  |[fill=blue]|   &  |[fill=blue]|  &  |[fill=blue]|  & 
|[fill=blue]|  &  |[fill=blue]| & |[fill=blue]| \\
|[fill=green]| & |[fill=green]|  &  |[fill=green]|   &  |[fill=green]|  &  |[fill=green]|  &  |[fill=green]|  &  |[fill=green]| & |[fill=green]| \\
|[fill=gray]|  &  |[fill=gray]| & |[fill=gray]| & |[fill=gray]|  &  |[fill=gray]| & |[fill=gray]| &|[fill=gray]|  &  |[fill=gray]|  \\
|[fill=cyan]| & |[fill=cyan]|  &  |[fill=cyan]|   &  |[fill=cyan]|  &  |[fill=cyan]|  &  |[fill=cyan]|  &  |[fill=cyan]| & |[fill=cyan]| \\
|[fill=brown]| & |[fill=brown]|  &  |[fill=brown]|   &  |[fill=brown]|  &  |[fill=brown]|  &  |[fill=brown]|  &  |[fill=brown]| & |[fill=brown]| \\
};

\draw[->] (1.5,0) -- (3.5,0); 

\node[] (nel_label) at (-2.0, 0.0) {NE};

\draw[-,black!30] (-1, -1.25) -- (6, -1.25);
\matrix[matrix of nodes, nodes={draw,minimum size=1mm}, nodes in empty cells,column sep=-\pgflinewidth,row sep=-\pgflinewidth] (el1) at (0,-2.25)
{
|[fill=red]| & |[fill=red]|  &  |[fill=red]|   &  |[fill=red]|  &  |[fill=red]|  &  |[fill=red]|  &  |[fill=red]| & |[fill=red]|  \\
|[fill=blue]| & |[fill=blue]|  &  |[fill=blue]|   &  |[fill=blue]|  &  |[fill=blue]|  & 
|[fill=blue]|  &  |[fill=blue]| & |[fill=blue]| \\
|[fill=cyan]| & |[fill=cyan]|  &  |[fill=cyan]|   &  |[fill=cyan]|  &  |[fill=cyan]|  &  |[fill=cyan]|  &  |[fill=cyan]| & |[fill=cyan]| \\
};

\node[] () at (5, -3.0) {$\vdots$};

\matrix[matrix of nodes, nodes={draw,minimum size=1mm}, nodes in empty cells,column sep=-\pgflinewidth,row sep=-\pgflinewidth] (el1) at (0,-4.0)
{
|[fill=gray]|  &  |[fill=gray]| & |[fill=gray]| & |[fill=gray]|  &  |[fill=gray]| & |[fill=gray]| &|[fill=gray]|  &  |[fill=gray]|  \\
|[fill=cyan]| & |[fill=cyan]|  &  |[fill=cyan]|   &  |[fill=cyan]|  &  |[fill=cyan]|  &  |[fill=cyan]|  &  |[fill=cyan]| & |[fill=cyan]| \\
|[fill=brown]| & |[fill=brown]|  &  |[fill=brown]|   &  |[fill=brown]|  &  |[fill=brown]|  &  |[fill=brown]|  &  |[fill=brown]| & |[fill=brown]| \\
};

\matrix[matrix of nodes, nodes={draw,minimum size=1mm}, nodes in empty cells,column sep=-\pgflinewidth,row sep=-\pgflinewidth] (el1) at (5,-2.25)
{
|[fill=red]| & |[fill=red]|  &  |[fill=white]|   &  |[fill=red]|  &  |[fill=red]|  &  |[fill=red]|  &  |[fill=red]| & |[fill=red]|  \\
|[fill=blue]| & |[fill=blue]|  &  |[fill=blue]|   &  |[fill=blue]|  &  |[fill=blue]|  & 
|[fill=blue]|  &  |[fill=blue]| & |[fill=blue]| \\
|[fill=cyan]| & |[fill=cyan]|  &  |[fill=cyan]|   &  |[fill=cyan]|  &  |[fill=cyan]|  &  |[fill=cyan]|  &  |[fill=cyan]| & |[fill=cyan]| \\
};

\node[] () at (0, -3.0) {$\vdots$};

\matrix[matrix of nodes, nodes={draw,minimum size=1mm}, nodes in empty cells,column sep=-\pgflinewidth,row sep=-\pgflinewidth] (el1) at (5,-4.0)
{
|[fill=gray]|  &  |[fill=gray]| & |[fill=gray]| & |[fill=gray]|  &  |[fill=gray]| & |[fill=gray]| &|[fill=gray]|  &  |[fill=gray]|  \\
|[fill=cyan]| & |[fill=cyan]|  &  |[fill=cyan]|   &  |[fill=cyan]|  &  |[fill=cyan]|  &  |[fill=white]|  &  |[fill=cyan]| & |[fill=cyan]| \\
|[fill=brown]| & |[fill=brown]|  &  |[fill=brown]|   &  |[fill=brown]|  &  |[fill=brown]|  &  |[fill=brown]|  &  |[fill=brown]| & |[fill=brown]| \\
};

\draw[->] (1.5,-2.25) -- (3.5,-2.25); 
\draw[->] (1.5,-4) -- (3.5,-4); 

\node[] (el_label) at (-2.0, -3.0) {EP};
\node[] (el_episode_1) at (-1.35, -2.25) {$E_1$};
\node[] (el_episode_m) at (-1.35, -4.0) {$E_M$};
\draw[-, black!30] (-1, -4.75) -- (6, -4.75);

\matrix[matrix of nodes, nodes={draw,minimum size=1mm}, nodes in empty cells,column sep=-\pgflinewidth,row sep=-\pgflinewidth] (el1) at (-0.4,-6.0)
{
|[fill=red]| & |[fill=red]|  &  |[fill=red]|   &  |[fill=red]|  &  |[fill=red]|   \\
|[fill=blue]| & |[fill=blue]|  &  |[fill=blue]|   &  |[fill=blue]|  &  |[fill=blue]|  \\
|[fill=cyan]| & |[fill=cyan]|  &  |[fill=cyan]|   &  |[fill=cyan]|  &  |[fill=cyan]| \\
};
\matrix[matrix of nodes, nodes={draw,minimum size=1mm}, nodes in empty cells,column sep=-\pgflinewidth,row sep=-\pgflinewidth] (el1) at (0.8,-6.0)
{
 |[fill=red]|  &  |[fill=red]| & |[fill=red]|  \\
|[fill=blue]|  &  |[fill=blue]| & |[fill=blue]| \\
 |[fill=cyan]|  &  |[fill=cyan]| & |[fill=cyan]| \\
};

\node[] () at (0, -6.65) {$\vdots$};

\matrix[matrix of nodes, nodes={draw,minimum size=1mm}, nodes in empty cells,column sep=-\pgflinewidth,row sep=-\pgflinewidth] (el1) at (-0.4,-7.5)
{
|[fill=gray]|  &  |[fill=gray]| & |[fill=gray]| & |[fill=gray]|  &  |[fill=gray]|  \\
|[fill=cyan]| & |[fill=cyan]|  &  |[fill=cyan]|   &  |[fill=cyan]|  &  |[fill=cyan]|\\
|[fill=brown]| & |[fill=brown]|  &  |[fill=brown]|   &  |[fill=brown]|  &  |[fill=brown]|  \\
};
\matrix[matrix of nodes, nodes={draw,minimum size=1mm}, nodes in empty cells,column sep=-\pgflinewidth,row sep=-\pgflinewidth] (el1) at (0.8,-7.5)
{
 |[fill=gray]| & |[fill=gray]|  &  |[fill=gray]|  \\
 |[fill=cyan]| &  |[fill=cyan]| & |[fill=cyan]| \\
 |[fill=brown]| &  |[fill=brown]| & |[fill=brown]| \\
};

\matrix[matrix of nodes, nodes={draw,minimum size=1mm}, nodes in empty cells,column sep=-\pgflinewidth,row sep=-\pgflinewidth] (el1) at (4.6,-6.0)
{
|[fill=red]| & |[fill=red]|  &  |[fill=red]|   &  |[fill=red]|  &  |[fill=red]|   \\
|[fill=blue]| & |[fill=blue]|  &  |[fill=blue]|   &  |[fill=blue]|  &  |[fill=blue]|  \\
|[fill=cyan]| & |[fill=cyan]|  &  |[fill=cyan]|   &  |[fill=cyan]|  &  |[fill=cyan]| \\
};
\matrix[matrix of nodes, nodes={draw,minimum size=1mm}, nodes in empty cells,column sep=-\pgflinewidth,row sep=-\pgflinewidth] (el1) at (5.85,-6.0)
{
 |[fill=white]|  &  |[fill=white]| & |[fill=white]|  \\
|[fill=white]|  &  |[fill=white]| & |[fill=white]| \\
 |[fill=white]|  &  |[fill=white]| & |[fill=white]| \\
};

\node[] () at (5, -6.65) {$\vdots$};

\matrix[matrix of nodes, nodes={draw,minimum size=1mm}, nodes in empty cells,column sep=-\pgflinewidth,row sep=-\pgflinewidth] (el1) at (4.6,-7.5)
{
|[fill=gray]|  &  |[fill=gray]| & |[fill=gray]| & |[fill=gray]|  &  |[fill=gray]|  \\
|[fill=cyan]| & |[fill=cyan]|  &  |[fill=cyan]|   &  |[fill=cyan]|  &  |[fill=cyan]|\\
|[fill=brown]| & |[fill=brown]|  &  |[fill=brown]|   &  |[fill=brown]|  &  |[fill=brown]|  \\
};
\matrix[matrix of nodes, nodes={draw,minimum size=1mm}, nodes in empty cells,column sep=-\pgflinewidth,row sep=-\pgflinewidth] (el1) at (5.85,-7.5)
{
 |[fill=white]| & |[fill=white]|  &  |[fill=white]|  \\
 |[fill=white]| &  |[fill=white]| & |[fill=white]| \\
 |[fill=white]| &  |[fill=white]| & |[fill=white]| \\
};
\draw[->] (1.5,-6.0) -- (3.5,-6.0); 
\draw[->] (1.5,-7.5) -- (3.5,-7.5); 

\node[] (epsq_label) at (-1.75, -6.75) {EPSQ};
\node[] (episode_1) at (-1.35, -6.0) {$E_1$};
\node[] (episode_m) at (-1.35, -7.5) {$E_M$};

\node[] (episode_1) at (-0.5, -5.25) {support};
\node[] (episode_1) at (0.75, -5.25) {query};

%
\end{tikzpicture}
    }
    \caption{
    A illustration of training techniques, i.e., Non-episodic (\NEL), episodic (\EL) and episodic with a support and query split (\ELSQ).  
    Each color depicts a high-resource intent class. 
    Each row shows labeled utterances of high-resource intents. 
    White cells show example queries for which loss is computed. 
    In NE and EP training, the average loss over all cells are computed to update the parameters.
    }
    \label{fig:training}
\end{figure}
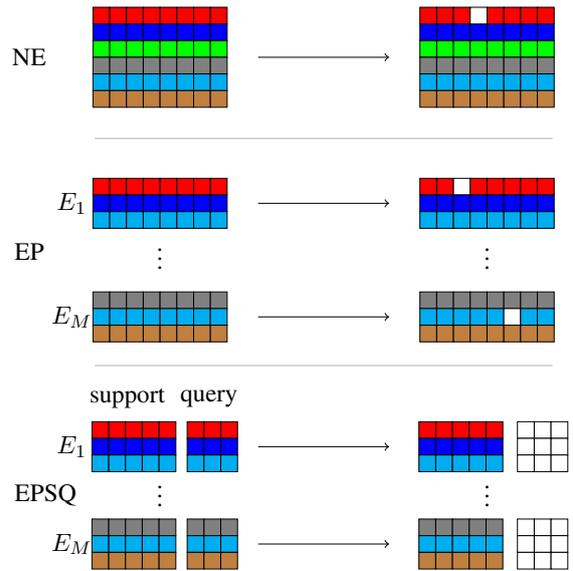

\subsection{Training Regimes} 
We empirically evaluate three training techniques (\cref{fig:training}) to train the aforementioned similarity functions: 
\emph{Non-Episodic Training (\NEL)}, 
\emph{Episodic Training (\EL)} and 
\emph{Episodic Training with Support and Query splits (\ELSQ)}. 
The training strategies rely on an identical loss function for each query. 

\paragraph{Loss per query sample.}
We use the loss function defined by \newcite{zhang-etal-2020-discriminative} for FSIC. 
In particular, we define a ground-truth binary vector $y_q$ for a query $q$ given a set of neighbours $C=\{ c_1,...,c_n \}$. 
If the query and its $i$-th neighbour belong to the same intent class, the corresponding label for the pair is $y_{q,i} = 1$, otherwise $y_{q,i} = 0$. 
Given such ground-truth label vector in consideration of the $n$ neighbours, $y_q = [y_{q,t}| t = 1,...,n]$ and similarity scores estimated by a model configuration for all pairs $s_q = [s_{q,t}| t = 1,...,n]$, we can compute the binary cross-entropy loss for the query $q$ as follows: 
\begin{multline}
    \label{eq:loss_q}
    l_q(y_q, s_q|C)  = \\  - \frac{1}{n} \sum_{t=1}^n \left[ y_{q,t}\log(s_{q,t}) + (1-y_{q,t})\log(1-s_{q,t}) \right] \texttt{.}
\end{multline}

\paragraph{\NEL.}
For the \NEL training, the classifier learns the semantic relation between all high-resource intent classes altogether.
Let $D$ represent a large set of utterances for high-resource intent classes. 
Therefore, we take each utterance in $D$ as a query $q$ and predict its label concerning the rest of the utterances as neighbors.   
More formally, we estimate the loss for the \NEL training as follows: 
\begin{equation}
    \mathcal{L} = \frac{1}{|D|} \sum_{q \in D} l_{q|D-q} (y_q, s_q)\texttt{,}
\end{equation}
where $l_q$ is the loss defined in Equation~\ref{eq:loss_q} between ground truth label vector $y_q$ and a vector of scores $s_q$ estimated by a similarity function.  
In particular, let $s_q = [ s_{q,1}, ..., s_{q,|D-1|} ] $ represent a vector of scores estimated by one of our configurations between the query $q$ and any other utterance in $D$. 

\paragraph{\EL.}
An episode is a set of utterances for several intent classes. 
An episode formulates an $N$-way intent classification task, where $N$ is the number of intent classes in the episode.  
The core idea behind meta-learning is to learn from a large set of high-resource intent classes by chunking the set into many episodes. 
These episodes are known as training episodes (a.k.a meta-training episodes).  
If set $\mathcal{I}$ denotes the intent labels of a benchmark corpus, any $N$ randomly selected intents from $\mathcal{I}$ can be used to construct an episode. 
Let's refer to these selected intents for episode $E$ by $\mathcal{I}_E$. 
Then, episode $E$ contains utterances $x$ whose intent labels $y$ are in $I_E$. 
It is worth noting that intent classes in training episodes may overlap to let a classifier learn the semantic relations between all intent labels of the benchmark. 
In \EL, we construct many episodes from high-resource intent classes. 
We define the following loss function: 
\begin{equation}
     \mathcal{L} = \frac{1}{M} \sum_{i=1}^{M} \frac{1}{|E_i|}\sum_{q \in E_i} l_{q|E_i} (y_q, s_q)\texttt{,}
\end{equation}
where $M$ is the number of episodes constructed from $D$, $E_i$ is the $i$th episode, $y_q$ is the ground-truth labels for the query given neighbors in the episode $E_i$, and $s_q$ is the similarity score estimated by a similarity function between the query and any neighbor in the episode.

\paragraph{\ELSQ.}
Intuitively, to imitate the few-shot setup, an episode can be split into two disjoint sets: a support and a query set.  
An episode's support set includes only a few utterances from each intent class in $I_E$.   
An episode's query set includes the rest of the utterances in the episode. 
A classifier should classify utterances in the query set using the utterances and intent labels in the support set.   
Given the $k$NN terminology, the support set is the set of neighbours and query set is a set of query utterances. 
Therefore, the main difference between \ELSQ and \EL is that the number of neighbours in \ELSQ  is limited to only a few examples of each intent. 
The loss function in \ELSQ is defined as follows: 
\begin{equation}
    \mathcal{L} = \frac{1}{M} \sum_{i=1}^{M} \frac{1}{|Q_i|}\sum_{q \in Q_i} l_{q|S_i} (y_q, s_q)\texttt{,}
\end{equation}
where $Q_i$ is the query set and $S_i$ is the support set of the $i$th episode.

\section{Experiments}
\label{sec:experiments}

We conduct our experiments in two different setups: (i) $N$-way $k$-shot and (ii) imbalanced classes in the support sets. 
The $N$-way $k$-shot setup refers to the typical few-shot learning setup, where the numbers of classes and examples per class are balanced.
In contrast, the imbalanced setup randomly defines the numbers of classes and examples, imitating the imbalance nature of most supervised benchmarks for intent classification. 
While arguably some utterances can be annotated to transform imbalanced episodes into balanced ones, imbalanced few-shot learning is still a huge practical challenge for various expensive domains, e.g., those that require experts for annotation \cite{krone-etal-2020-learning}. 

\begin{table}[t]
    \small
    \centering
    \begin{tabular}{lrrrc}
    \toprule
        \multirow{2}{*}{Dataset} 
        & 
        \multicolumn{3}{c}{\#classes}
        & \#shots
        \\
        & train & valid & test & per intent\\
        \midrule
        \multicolumn{4}{l}{\emph{Balanced}} \\
        \midrule
        Clinc (150) & 50 & 50 & 50 & \multirow{4}{*}{\{1,5\}} \\
        Banking (77) & 25 & 25 & 27 \\
        Liu (54) & 18 & 18 & 18 \\
        Hwu (64) & 23 & 16.4 & 24.6 \\
        \midrule
        \multicolumn{4}{l}{\emph{Imbalanced}} \\
        \midrule
        ATIS (19) & 5 & 7 & 7 & \multirow{3}{*}{variable} \\
        SNIPS (7) & 4 & - & 3 \\
        TOP (18) & 7 & 5 & 6 \\
    \bottomrule
    \end{tabular}
    \caption{Experimental datasets and their main statistics (number of distinct classes per data split, number of examples/shots per intent/class). 
    The numbers in parenthesis show the total number of intents datasets.
    For HWU64, each split’s number of classes varies at each run to ensure there is no cross-split domain, hence the decimal number.
    }
    \label{tab:data_statistics}
\end{table}

\def\arraystretch{1}
\begin{table*}[!t]
    \small
    \centering
    \begin{tabular}{lccccc|ccccc }
    \toprule
    & \multicolumn{5}{c|}{1-shot} & \multicolumn{5}{c}{5-shot}\\
    & Clinc & Banking & Liu & Hwu & Avg  & Clinc & Banking & Liu & Hwu & Avg  
    \\
    \midrule
    Random &  20.17 & 20.17 & 20.17 & 20.17 & 20.17 & 19.71 & 19.71 & 19.71 & 19.71 & 19.71 \\
    \BE(fixed)+NP  &  30.88 & 27.75 & 30.83 & 29.49 & 29.74 & 48.57 & 38.01 & 45.79 & 41.15 & 43.38 \\
    ProtoNet & 94.29 & 82.20 & 80.06 & 74.37 & 82.73 & 98.10 & 91.57 & 89.62 & 86.48 & 91.44 \\
    \midrule
    \CE+\Pa &&&&&\\
    \hspace{3mm} \NEL & 58.45 & 48.88 & 48.98 & 50.12 & 51.61 & 66.93 & 64.46 & 55.83 & 59.35 & 61.64 \\
    \hspace{3mm} \EL & 93.60 & 79.46 & 77.36 & 72.13 & 80.64 & 98.26 & \textbf{92.38} & \textbf{88.33} & 84.43 & \textbf{90.85} \\
    \hspace{3mm} \ELSQ & \textbf{94.65} & \textbf{79.82} & \textbf{78.13} & \textbf{72.64} & \textbf{81.31} & \textbf{98.49} & 92.15 & 88.18 & \textbf{84.59} & \textbf{90.85} \\
    \midrule
    \BE+\Pa &&&&&\\
    \hspace{3mm} \NEL & 79.48 & 60.26 & 59.15 & 52.04 & 62.73 & 88.04 & 70.28 & 70.49 & 60.47 & 72.32 \\
    \hspace{3mm} \EL & 82.66 & 66.43 & 59.76 & 50.53 & 64.85 &  92.87 & 77.99 & 70.60 & 61.18 & 75.66 \\
    \hspace{3mm} \ELSQ &83.26 & 66.53 & 60.41 & 51.40 & 65.40 & 92.51 & 78.59 & 70.82 & 64.13 & 76.51 \\
    \midrule
    \BE+NP &&&&&\\
    \hspace{3mm} \NEL & 58.04 & 45.24 & 53.18 & 42.57 & 49.76 & 78.10 & 68.57 & 61.52 & 54.86 & 65.76 \\
    \hspace{3mm} \EL & 67.58 & 52.85 & 52.39 & 41.73 & 53.64 & 76.28 & 67.69 & 63.33 & 51.37 & 64.67 \\
    \hspace{3mm} \ELSQ & 67.80 & 53.83 & 53.17 & 40.96 & 53.94  & 81.31 & 64.58 & 65.32 & 50.41 & 65.41 \\
    \bottomrule
    \end{tabular}
    \caption{\label{tab:main_results} \textbf{BERT-based} results of experimental methods on 4 datasets with different numbers of examples per class (1 or 5 examples per class) setting. }
\end{table*}

\paragraph{Datasets, splits, and episodes.}
\cref{tab:data_statistics} summarizes the main statistics (e.g., the number of classes per data split for each datasets) of the datasets and their splits as we use in our experiments. 
For the balanced $N$-way $k$-shot setup, we use Clinc~\cite{larson-etal-2019-evaluation}, Banking~\cite{casanueva-etal-2020-efficient}, and Hwu~\cite{liu2021benchmarking} from DialoGLUE~\cite{MehriDialoGLUE2020} as well as Liu~\cite{liu2021benchmarking}. 
For the sake of fair comparisons, we use the exact splits and episodes as used by  \newcite{dopierre-etal-2021-protaugment} for FSIC.  
For 5 folds, we randomly split intents in each dataset into three sets to construct train, valid and test episodes. 
We then generate $5$-way $k$-shot episodes for each split in each fold, where $k\in \{1,5\}$. 
For the imbalanced setup, we use ATIS~\cite{hemphill-etal-1990-atis}, SNIPS~\cite{coucke2018snips}, and TOP~\cite{gupta-etal-2018-semantic}.  
We follow \newcite{krone-etal-2020-learning} to construct episodes on these datasets.
We refer interested readers in details of the episode construction to Appendix. 

\paragraph{Settings.}
We use BERT-based-uncased and SIMCSE as language models. 
We use AdamW optimizer \cite{loshchilov-hutter-2019-decoupled} with a learning rate of $2e-5$. 
Both batch size and maximum sequence length are set to $64$.
See the Appendix for the full list of hyperparameters. 
For experiments on balanced datasets, we conduct the experiments over 5 folds.
For each fold, we train a FSIC classifier for a maximum of 10,000 $5$-way $K$-shots episodes.
We evaluate the model on the validation set after every $100$ updates, and stop the training if validation performance does not improve over 5 consecutive evaluation steps. 
To alleviate the impact of random selection of few-shot samples, we report the average performance of a classifier for 600 episodes, compatible with \newcite{dopierre-etal-2021-protaugment}. 
For the experiments on the imbalanced datasets, similar to \newcite{krone-etal-2020-learning}, we conduct the experiments over 1 fold due to the limited number of intents.  
The number of episodes in train, valid, and test splits of these datasets is respectively as follow, ATIS (1377, 213, 119), SNIPS (240, -, 210) and TOP (10095, 1286, 292). 
The average number of shots per intent used in episodes of ATIS, SNIPS, and TOP is about 4, 5, and 4, respectively (see Appendix for details). 
For both $N$-way $k$-shot and imbalanced setups, the number of examples in query sets is identical for all intents in the query sets. 
So we report the accuracy metric for all experiments average over all runs and folds.

\paragraph{Models compared.}
Alongside the results of the model configurations (\cref{sec:method}), we report the results of the following FSIC methods to put our results in context. 
\textbf{Random} assigns a random intent class from the support set to each query utterance. 
\textbf{\text{BE (fixed)+NP}} represents a generic configuration for the usage of majority of LM-based FSIC baselines, e.g., ConvBERT~\cite{mehri-eric-2021-example},  TOD-BERT~\cite{wu-etal-2020-tod}, and DNNC-BERT~\cite{zhang-etal-2020-discriminative}, etc.
These methods use pretrained BERT and further fine-tune for other NLP tasks (e.g., NLI) or conversational datasets. 
\textbf{ProtoNet \cite{dopierre-etal-2021-protaugment}} is inspired by prototypical network method \cite{snell-et-al-2017}, which has been shown to achieve the state-of-the-art accuracy among meta-learning methods for few-shot learning tasks including FSIC \cite{krone-etal-2020-learning}.  
We remark that this method is not based on instance similarity. 
In fact, it encodes an intent class by a prototype vector, which is the mean of vector representations of a few utterances given for the intent. 
Given an episode the prototype vector of  each intent is computed and  then the probability of intent classes is estimated based on the distances between an utterance vector and intent prototypes.

\def\arraystretch{1}
\begin{table*}[t]
    \small
    \centering
    \begin{tabular}{lccccc|ccccc }
    \toprule
    & \multicolumn{5}{c|}{1-shot} & \multicolumn{5}{c}{5-shot}\\
    & Clinc & Banking & Liu & Hwu & Avg  & Clinic & Banking & Liu & Hwu & Avg  
    \\
    \midrule
    \BE (fixed) +\Me &  91.33 & 75.48 & 78.75 & 74.58 & 80.03 &97.89 & 90.33 & 89.61 & 86.93 & 91.19 \\
    \midrule
    \CE+\Pa &&&&&\\
    \hspace{3mm} \NEL & 60.51 & 54.87 & 49.99 & 46.41 & 52.95 & 78.33 & 72.71 & 68.66 & 67.99 & 71.92 \\
    \hspace{3mm} \EL & 94.33 & 83.64 & 79.24 & 77.03 & 83.56 &  \textbf{98.80} & \textbf{94.22} & \textbf{90.13} & \textbf{88.54} & \textbf{92.92} \\
    \hspace{3mm} \ELSQ &  \textbf{95.01} & \textbf{83.83} & 79.40 & \textbf{77.49} & \textbf{83.93} & 98.77 & 94.04 & 90.10 & 88.40 & 92.83 \\
    \midrule
    \BE+\Pa &&&&&\\
    \hspace{3mm} \NEL &  90.69 & 76.21 & 68.76 & 66.05 & 75.43 &96.71 & 88.12 & 80.76 & 79.26 & 86.21 \\
    \hspace{3mm} \EL &  90.93 & 76.81 & 71.32 & 65.72 & 76.19 &96.74 & 88.18 & 83.83 & 80.64 & 87.35 \\
    \hspace{3mm} \ELSQ &  90.95 & 76.43 & 71.33 & 65.71 & 76.11 &96.83 & 87.95 & 84.10 & 80.90 & 87.45 \\
    \midrule
    \BE+NP &&&&&\\
    \hspace{3mm} \NEL &  93.69 & 81.60 & 79.51 & 75.54 & 82.58 &98.08 & 91.56 & 89.61 & 87.82 & 91.77 \\
    \hspace{3mm} \EL &  93.24 & 80.15 & 79.82 & 76.49 & 82.43 &98.01 & 91.91 & 89.77 & 87.62 & 91.83 \\
    \hspace{3mm} \ELSQ &  93.44 & 80.46 & \textbf{80.21} & 76.68 & 82.70 &98.02 & 91.95 & 89.83 & 87.65 & 91.86 \\
    \bottomrule
    \end{tabular}
    \caption{\label{tab:results_simcse} \textbf{SimCSE-based} results of experimental methods on 4 datasets with different numbers of examples per class (1 or 5 examples per class) setting. }
\end{table*}

\def\arraystretch{1}
\begin{table}[!b]
    \small
    \centering
    \begin{tabular}{lcccc}
    \toprule
      & ATIS       & SNIPS       &TOP   & Avg \\
                                                                                   
    \midrule
    Random          & 21.34                & 33.70                & 23.99     & 26.34    \\
    \BE (fixed) + \Me  & 53.80                & 51.62                & 33.03    &  46.15       \\
    \midrule
    \CE + \Pa   &                      &                      &                 \\
    \hspace{3mm}\NEL       & 62.86                & 65.03                & 49.41     & 59.10      \\
    \hspace{3mm}\EL     & \textbf{79.71}       & \textbf{93.94}       & \textbf{68.04} & \textbf{80.56} \\
    \hspace{3mm}\ELSQ                 & 71.58      & 92.98       & 62.84 & 75.80\\
    \midrule
    \BE + \Pa   &                 \\
    \hspace{3mm}\NEL    & 42.91                & 80.22                & 53.48     & 58.87      \\
    \hspace{3mm}\EL   & 69.52                & 60.54                & 51.46     & 60.51      \\
    \hspace{3mm}\ELSQ   & 66.44                & 62.21                & 56.05   & 61.57        \\
    \midrule
    \BE + \Me  &                      &                      &                 \\
    \hspace{3mm}\NEL                                         & 65.86                & 77.92                & 45.52   & 63.10        \\
    \hspace{3mm}\EL                                                           & 65.09                & 79.16                & 47.85     & 64.03      \\
    \hspace{3mm}\ELSQ                                                        & 55.67                & 80.08                & 42.97   & 59.57        \\
    \bottomrule
    \end{tabular}
    \caption{Results of experimental methods on three datasets with imbalanced classes and shots setting. }
    \label{tab:imbalanced_results}
\end{table}

\section{Results and Discussion}
We compare the the configurations described in (\cref{sec:method}) and baselines (\cref{sec:experiments}) for balanced and imbalanced FSIC setups using BERT and SimCSE, which is the state-of-the-art sentence encoder.  
The main experimental findings are 
\begin{itemize}
    \item The cross-encoder architecture with parameterized similarity scoring function and episodic meta-learning consistently yields the best FSIC accuracy. 
    \item Episodic training yields a more robust FSIC classifier than non-episodic one for most examined datasets. 
    \item Splitting episodes to support and query sets is not a must for episodic training. 
\end{itemize}

\subsection{Balanced FSIC}
\cref{tab:main_results} shows the accuracy of the examined models using BERT when 1-shot and 5-shots are given.  
The accuracy of the examined few-shot classifiers is consistently superior to the ``\BE (fixed)+\Me'' baseline.
This shows that fine-tuning BERT high-resource intent classes by the training techniques improves BERT's generalization capability to unseen intents. 
For both 1-shot and 5-shots, \CE + \Pa trained with either of episodic training strategies, achieves a higher accuracy (by about 29\% on average) than when it is trained with \NEL, and  performs near the ProtoNet method.   
This shows that the examined episodic training is more effective than the non-episodic training in increasing the robustness of the FSIC models with the cross-encoder architecture. 
\BE + \Pa trained with episodic training achieves about 2\% higher accuracy for 1-shot and 3\% for 5-shots than when it is trained with \NEL, confirming the effectiveness of episodic training. 
For \BE + \Me, EP results in 3.8\% higher accuracy than \NEL for 1-shot and a similar accuracy to \NEL for 5-shots.
Overall, using the BERT encoder, episodic training is more effective than non-episodic training for the examined combinations of pairwise encoders and similarity functions.

\ELSQ achieves a similar accuracy to \EL (less than 1\% difference) for all \CE+\Pa, \BE+\Pa, and \BE+\Me on average. 
We induce that splitting utterances of an episode into a support and a query set does not impact the capability of the classifier to generalize to unseen intents. 
This observation is strongly important because all episodic learning FSIC  approaches~\cite{dopierre-etal-2021-protaugment, krone-etal-2020-learning} use the support-query split assumption. 

We observe that by increasing the number of labled utterances from 1-shot to 5-shots for each intent, the FSIC performance increases. \BE+\Me trained with \NEL obtains the highest improvement by 16.01\%, and the other few-shot classifiers achieve about 10.5\% improvement on average. 
This makes sense because more training shots help the method to learn more discriminative representations of intent classes.  

To study if further pre-training or intermediate fine-tuning \cite{phang2018sentence} close the gap among the explored analysis dimensions, we substitute BERT with SimCSE (results in \cref{tab:results_simcse}). 
Our three main findings hold in this experiment as well. 
Importantly, only \text{\CE + \Me} trained with episodic training strategies outperforms ``\BE (fixed) + \Me'' by at least 3.5\% for 1-shot and 1.6\% for 5-shots on average.  
This reveals the effectiveness of cross-encoders  and episodic training for FSIC. 
However, different from the BERT-based results, not all the examined configurations beat \text{``\BE (fixed) + \Me''}, which might be due to SimCSE's enriched vectors. 

\subsection{Imbalanced FSIC}
\cref{tab:imbalanced_results} shows the results on the three imbalanced datasets.
\CE + \Pa with \EL substantially achieves higher accuracy compared with its counterparts, defining this never-explored FSIC method as a new effective method for imbalanced FSIC. 
On average, episodic training (EP) outperforms non-episodic (NE) training strategy. 
The \CE + \Pa and \BE + NP configurations achieve a higher accuracy when trained by EP compared with when trained with EPSQ, showing that the common belief in meta-learning that episodes should be split into support and query set does not hold for FSIC. 
The only exception is BE + PA, which might be because of over parameterzied parametric function which is used to map the concatenation of two sentence vector to a similarity score. 
These results are compatible with our main findings in balanced FSIC.


\section{Conclusions}
We shed light on factors that contribute to performance of few-shot intent classification (FSIC), which is a crucial task in modular dialogue systems. 
We categorize FSIC approaches across three essential dimensions: 
(1) the cross-encoder vs bi-encoder encoding architectures;
(2) the parametric (i.e., trainable) vs non-parametric scoring functions (i.e. cosine similarity); and 
(3) episodic vs non-episodic training.  
We experiment on seven standard FSIC datasets. We show that the cross-encoder architecture (with parameterized similarity scoring function) trained with episodic training consistently yields the best performance for few-shot intent classification. 
Episodic training perform better than non-episodic on most examined datasets. 
The hypothesis in meta-learning to split episodes to support and query sets is not a crucial factor for episodic training of FSIC.

\section*{Limitations and Ethical Concerns.}
In this paper, we shed light to few-shot intent classification tasks in modular (task-oriented) dialogue systems. Dialog systems, given their direct interaction with human users, must be devoid of any negative stereotypes and must not exhibit any behaviour that could be potentially harmful to humans. That said, our work does not address the generation component of dialog systems, but merely the intent classification. As such, we do not believe it raises any ethical concerns.    

The main limitation of the work -- conditioned primarily by the available computational resources -- is the scope of our empirical comparison: we focus on FSIC methods that subscribe to pairwise similarity scoring of utterances and nearest neighbours inference. While this subsumes much of the best performing approaches in the literature, there is a fair body of recent work that does not fall in this group.  
Another limitation of the work is the monolingual focus on English only. We intend to extend our work to cross-lingual transfer to other languages, for which fewer labeled intent classification datasets exist. 


\bibliography{anthology,references}
\bibliographystyle{acl_natbib}

\end{document}